\documentclass[conference]{IEEEtran}
\IEEEoverridecommandlockouts

\usepackage{xcolor}
\usepackage[mathscr]{euscript}
\usepackage[ruled,linesnumbered,vlined,shortend]{algorithm2e}
\usepackage{subfigure}
\usepackage{cite}
\usepackage{amsmath,amssymb,amsfonts}
\usepackage{algorithmic}
\usepackage{graphicx}
\usepackage{textcomp}
\usepackage{xcolor}
\def\BibTeX{{\rm B\kern-.05em{\sc i\kern-.025em b}\kern-.08em
    T\kern-.1667em\lower.7ex\hbox{E}\kern-.125emX}}
\begin{document}

\title{Global and Local Structure Learning for Sparse Tensor Completion
}

\author{\IEEEauthorblockN{Dawon Ahn}
\IEEEauthorblockA{\textit{Computer Science and Engineering} \\
\textit{University of California, Riverside}\\
Riverside, USA \\
dahn017@ucr.edu}
\and
\IEEEauthorblockN{Evangelos E. Papalexakis}
\IEEEauthorblockA{\textit{Computer Science and Engineering} \\
\textit{University of California, Riverside}\\
Riverside, USA \\
epapalex@cs.ucr.edu}
}

\maketitle
\newtheorem{defn}{Definition}
\newtheorem{lem}{Lemma}
\newtheorem{theo}{Theorem}
\newenvironment{customlemma}[1]
{\renewcommand\theinnercustomlemma{#1}\innercustomlemma}
{\endinnercustomlemma}

\newcommand{\argmin}[1]{\underset{#1}{\operatorname{arg}\,\operatorname{min}}\;}
\newcommand{\T}[1]{\boldsymbol{\mathscr{#1}}}
\newcommand{\tensor}[1]{\boldsymbol{\mathscr{#1}}}
\newcommand{\mat}[1]{\mathbf{#1}}
\newcommand{\vect}[1]{\mathbf{#1}}

\newcommand{\A}[1]{\mat{A}^{(#1)}}
\renewcommand{\a}[1]{\vect{a}^{(#1)}}
\newcommand{\R}[0]{\mat{R}}
\newcommand{\D}[0]{\mat{D}}

\newcommand{\reminder}[1]{{\textsf{\textcolor{red}  {[#1]}}}}

\newtheorem{algo}{Algorithm}

\newcommand{\blue}[1]{{\color{blue} #1}}
\newcommand{\red}[1]{{\color{red} #1}}
\newcommand{\orange}[1]{{\color{orange} #1}}
\newcommand{\green}[1]{{\color{teal} #1}}
\newcommand{\hide}[1]{}

\newcommand*{\QEDA}{\hfill\ensuremath{\blacksquare}}
\newcommand*{\QEDB}{\hfill\ensuremath{\square}}

\newcommand*\circled[1]{\tikz[baseline=(char.base)]{
  \node[shape=circle,draw,inner sep=0.5pt] (char) {#1};}}
\newcommand{\oast}[0]{\circled{$\ast$}}

\newcommand{\method}[0]{\textsc{TGL}\xspace}
\newcommand{\fullmethod}[0]{Tensor Decomposition Learning Global and Local Structures}%

\begin{abstract}
\textit{How can we accurately complete tensors by learning relationships of dimensions along each mode?}
Tensor completion, a widely studied problem,  is to predict missing entries in incomplete tensors.
Tensor decomposition methods, fundamental tensor analysis tools, have been actively developed to solve tensor completion tasks. 
However, standard tensor decomposition models have not been designed to learn relationships of dimensions along each mode, which limits to accurate tensor completion.
Also, previously developed tensor decomposition models have required prior knowledge between relations within dimensions to model the relations, expensive to obtain.
This paper proposes \method (\fullmethod) to accurately predict missing entries in tensors.
\method reconstructs a tensor with factor matrices which learn local structures with GNN without prior knowledges.
Extensive experiments are conducted to evaluate \method with baselines and datasets.
\end{abstract}

\maketitle

\section{Introduction} \label{intro}

\textit{
Given an incomplete tensor, 
how can we predict missing values learning relations within dimensions?
}

A tensor is a natural way to represent multi-dimensional data.
For example, 3-mode tensor (user, movie, time) represents movie rating data where users give rates to movies at specific times.
The goal of tensor completion is to impute missing entries in incomplete tensors which are partially observed.
The tensor completion problems have been actively studied 
	across diverse domain of applications such as recommender system~\cite{karatzoglou2010multiverse,song2019tensor,liu2019costco},
		anomaly detection~\cite{kwon2021slicenstitch}, computer vision~\cite{vasilescu2002multilinear}, and air quality analysis~\cite{ahn2022time}.

Tensor decomposition methods have played a key role in tensor analysis
such that they have significantly been developed to solve the completion tasks.
CP (CANDECOMP/PARAFAC)~\cite{harshman1970foundations,kiers2000towards} factorization is one of the most widely used tensor factorization models due to its simplicity and interpretability,
	which factorizes a tensor into a set of factor matrices and a core tensor which is restricted to be diagonal.		
However, the design of standard CP decomposition model does not consider explicitly relations between dimensions,
which degrades the performance in missing entries prediction. 

Recently,  various types of tensor decomposition methods have been proposed to leverage
	prior knowledges, such as temporal knowledge, correlations between dimensions, as a
regularization into loss function~\cite{yu2016temporal, ahn2022time}.
However, 
prior knowledges are expensive to obtain for general datasets.

In this paper, we propose \method (\fullmethod) to learn relationships between dimensions without prior knowledges.
We leverage Graph Neural Networks (GNNs) to learn local structures.
\method initially produces representations with a standard CP decomposition
and then generates relation matrices with the representations.
With representation and relation matrices, \method updates representations via GNNs for each mode.
We show \method effectively learn relationships and achieve comparable accuracy compared to existing baselines in tensor completion.


\section{Proposed Method} \label{method}

In this study, 
we utilized both a CP decomposition method and GNNs
to learn relations within dimensions of each mode for incomplete tensors without prior knowledges. 
In each iteration of training,
we generate K-nearest neighborhoods (KNN) graphs with the current factor matrices for each mode.
Then we update factor matrices using KNN graphs and GNNs.
With these newly updated factor matrices, we reconstruct a tensor to jointly train factor matrices and GNNs.

We generate graphs $\mat{R}_A, \mat{R}_B$ and $\mat{R}_C$ with factor matrices $\mat{A}$, $\mat{B}$, and $\mat{C}$ by computing pairwise cosine similarity for each mode.
A graph $G$ contains a set of nodes $V (|V| = N) $, a set of edges $E$, and an adjacency matrix $\R \in \mathbb{R}^{N\times N}$ as $G = (V, E)$.
Given the number of layers $l (1 \leq l \leq L)$, 
$l+1$th hidden convolutional layer of GNNs takes feature matrix of the $l$th layer and the adjacency matrix 
$$
\mat{H}^{(l+1)} = f(\mat{H}^{(l)}, \R) = \sigma(\hat{\R}\mat{H}^{(l)}\mat{W}^{(l)})
$$
where $\hat{\R} = \tilde{\D}^{-\frac{1}{2}}\tilde{\R}\tilde{\D}^{\frac{1}{2}}$ is a symmetric normalization of the self-connections added adjacency matrix 
$\tilde{\R} = \R + \mat{I}$, $\tilde{\D}$ is the diagonal node degree matrix with $\tilde{d}_{ii} = \sum_{j}\tilde{r}_{ij}$, $\mat{W}^{(l)} \in \mathbb{R}^{d_l \times d_{l+1}}$ is a trainable weight matrix.
$\sigma$ is a non-linear activation function.
%
Note that initial feature matrix $\mat{H}^{(0)}$ is the factor matrix obtained.

Given factor matrices $\mat{A}, \mat{B}$, and $\mat{C}$, 
relation matrices $\mat{R}_A, \mat{R}_B$ and $\mat{R}_C$, and 
GNN models $\{f_1, f_2, f_3\}$ for each mode, 
the new factor matrices are obtained by minimizing the following loss function:
	\begin{align}
    	\mathcal{L}\left(\mat{A}, \mat{B}, \mat{C}\right)
    		& = \sum_{\forall \alpha \in \Omega}{ \left({x}_{\alpha}-\sum_{r=1}^{R} \tilde{a}_{ir} \tilde{b}_{jr} \tilde{c}_{kr}\right)^{2}}
    		\label{eq:GNN_comp:obs}
    \end{align}
	where  $\tilde{a}_{ir}$, $\tilde{b}_{jr}$, and $\tilde{c}_{kr}$ is the $(i, r)$th , $(j, r)$th , and $(k, r)$th entry of
	$\tilde{\mat{A}} = f_1(\mat{A}, \mat{R}_{\mat{A}})$,
	$\tilde{\mat{B}} = f_2(\mat{B}, \mat{R}_{\mat{B}})$, and
	$\tilde{\mat{C}} = f_3(\mat{C}, \mat{R}_{\mat{C}})$.
I will explain details of network hyper parameters in Section~\ref{experiments}.

\section{Experiments} \label{experiments}

We perform experiments to evaluate the accuracy of \method compared to baselines in tensor completion.
\subsection{Experiments Setting}
\subsubsection{Datasets}
We evaluate the performance of \method and baselines on three real-world datasets.
\textbf{Yelp}~\cite{asghar2016yelp} is
	a business rating tensor with (user, business, time; rating) which includes
	70,817 movies, 15,579 users, 108 year-month time duration, and 100,0836 ratings.
	We sample each 5,000 in user and business such that the statistics of dataset used is
	(5,000, 5,000, 108) since the proposed model can not handle too large dataset.
\textbf{BBC-News} is a BBC news article tensor with (word, word, articles; co-occurrences) which includes
	100 words and 400 articles.
We split each data into training, validation, and test sets with the ratio of 8:1:1 considering only observable entries and
use validation set for early stopping.

\subsubsection{Baselines}
There are two baselines: 
CPD~\cite{shin2016fully} is a standard CP decomposition model considering only nonzeros with gradient descent optimization.
CoSTCo~\cite{liu2019costco} is a state-of-the-art tensor completion method exploiting Convolutional Neural Networks.

\subsubsection{Metrics}
We use NRE (Normalized Reconstruction Error) to evaluate the accuracy of tensor completion, which is defined as follows.
$$\text{NRE}= \frac{\sqrt{\sum_{\forall\alpha \in \Omega}{\left(x_{\alpha}-\hat{x}_{\alpha}\right)^{2}}}}{\sqrt{\sum_{\forall\alpha \in \Omega}{x_{\alpha}^2}}}.
$$

 \begin{figure}[ht]
	\centering
	{
	\subfigure[Yelp]{
        \includegraphics[width=0.8\linewidth]{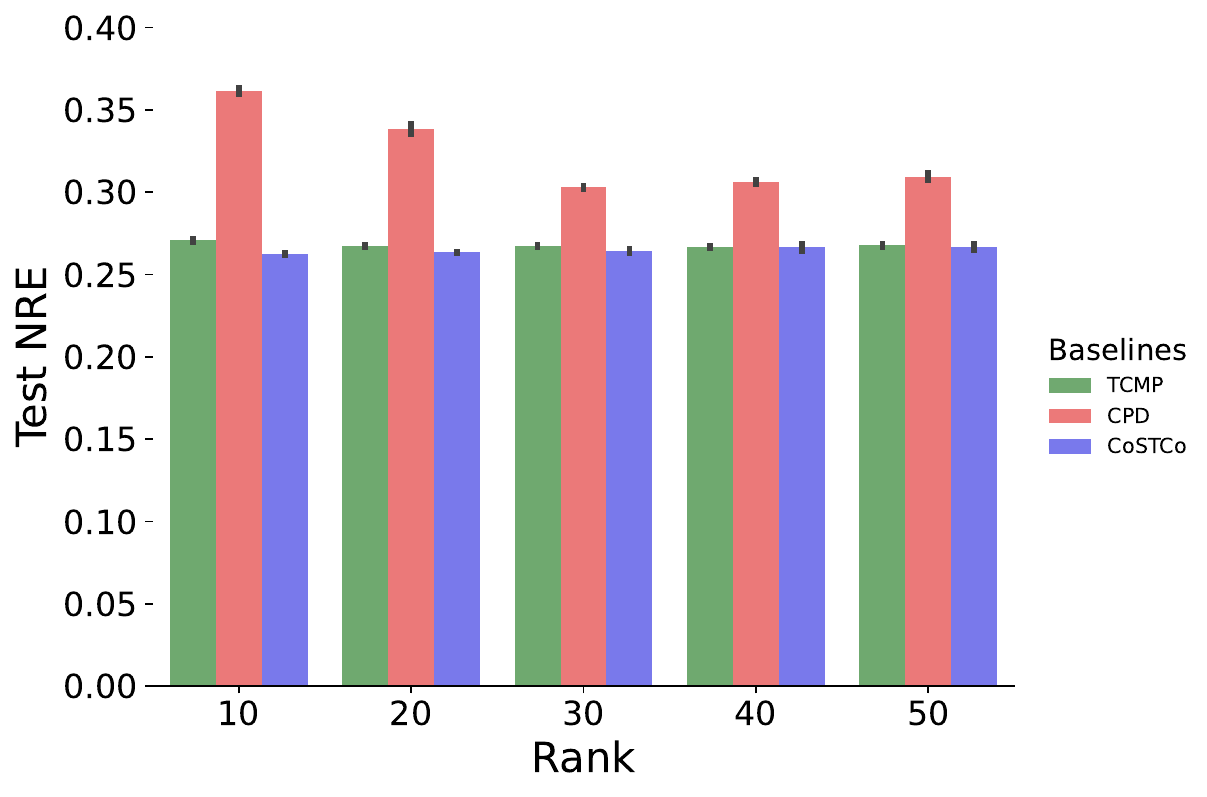}
    }
	\subfigure[BBC-News]{
        \includegraphics[width=0.8\linewidth]{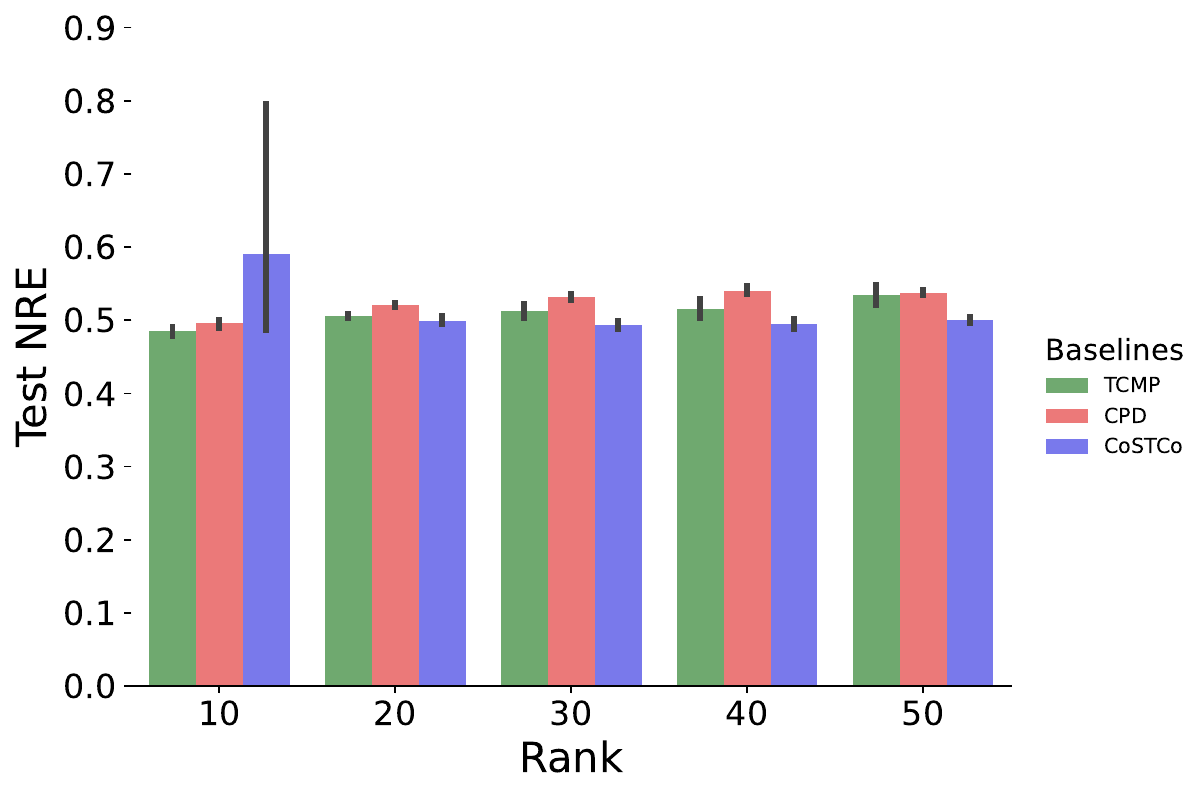}
    }
    }
    \caption{ \label{fig:perform}
    Comparison of test NRE of baselines on Yelp and BBC-News dataset with different size of ranks.}
\end{figure}

\subsection{Tensor Completion Accuracy} \label{sec:exp:accuracy}
Figure~\ref{fig:perform} show the performance of \method and baselines.
	with regard to test NRE.
\method shows the second best method
while CPD using the standard CP decomposition method shows large errors.
CoSTCo shows the lowest NRE among all methods
Even if \method shows better performance than the standard model, 
one potential reason that \method shows limited performance is that
when training GNNs, \method gives relation matrices and factor matrices which can have 
redundant information since relation matrices are created based on factor matrices.


\section{Discussion \& Conclusions } \label{conclusion}
In this paper, we proposed \method, 
a tensor decomposition assisted with GNNs to learn global and local structures of a sparse tensor.
Experiment results show that
\method outperforms CPD and shows comparable performance as CoSTCo.
As mentioned in Experiments section, 
a graph generated from \method might give redundant information to GNNs 
since relation matrices are made from factor matrices.
Therefore,  we try to make distinct graph structures from node features for GNNs for the future works.

\section*{Acknowledgments}
Research was supported by the National Science Foundation under CAREER grant no. IIS 2046086,  by the Agriculture and Food Research Initiative Competitive Grant no. 2020-69012-31914 from the USDA National Institute of Food and Agriculture, and by the University Transportation Center for Railway Safety (UTCRS) at UTRGV through the USDOT UTC Program under Grant No. 69A3552348340.
\bibliographystyle{IEEEtran}
\bibliography{paper.bib}
\end{document}